%% file: main.tex
\documentclass[11pt]{article}

\usepackage[final]{acl}

\usepackage{times}
\usepackage{latexsym}
\usepackage[T1]{fontenc}
\usepackage[utf8]{inputenc}
\usepackage{microtype}
\usepackage{inconsolata}
\usepackage{graphicx}
\usepackage{booktabs}
\usepackage{amsmath}
\usepackage{multirow}
\usepackage{xcolor}
\usepackage{array}
\usepackage{adjustbox}
\usepackage{makecell}
\usepackage{tikz}
\usetikzlibrary{arrows.meta,positioning,fit,backgrounds,shapes.misc}
\usepackage{listings}

\graphicspath{{figures/}}

\newcolumntype{L}[1]{>{\raggedright\arraybackslash}p{#1}}

\title{RuleChef: Grounding LLM Task Knowledge in Human-Editable Rules}

\author{Adam Kovacs \\
  KR Labs \\
  \texttt{kovacs@krlabs.eu}}

\author{
  \textbf{Ádám Kovács\textsuperscript{1}},
  \textbf{Nadia Verdha\textsuperscript{2}},
  \textbf{Gábor Recski\textsuperscript{1,2}}
\\
  \textsuperscript{1}KR Labs,
  \textsuperscript{2}TU Wien
\\
\small{
    \textbf{Correspondence:} \href{mailto:kovacs@krlabs.eu}{kovacs@krlabs.eu}
}
}

\begin{document}
\maketitle

\begin{abstract}
    We present RuleChef, a framework that uses large language models (LLMs) to generate executable
    rules for NLP tasks such as text classification, Named Entity Recognition (NER), or relation
    extraction. Rules are generated based on a task description and a set of labeled examples, then
    they are iteratively improved based both on additional examples and on human feedback over
    existing rules. RuleChef can also be used to bootstrap rules using the observed input-output
    pairs from any existing model for a given task. LLMs are used only at learning time,
    synthesizing rules and iteratively patching them based on failures measured on a held-out
    split. The result of this process is a fast, deterministic, and inspectable rule system.
    Preliminary evaluation is performed on
    both classification and NER tasks. 
    We release RuleChef as open-source software under an Apache 2.0
license\footnote{\url{https://github.com/KRLabsOrg/rulechef}}.
\end{abstract}

\section{Introduction}
\label{sec:intro}

Early NLP systems were predominantly rule-based: hand-crafted patterns and template grammars
provided transparent, deterministic models that could be inspected, debugged, and explained to end
users~\citep{chiticariu2013rule,valenzuela2020odinson,kovacs2022potato}.  The shift to
feature-based classifiers and then to pretrained transformers~\citep{devlin2019bert} and large
language models (LLMs) brought substantial gains in coverage and flexibility, but also moved decision
logic into latent parameters that are difficult to audit or edit directly~\citep{lipton2018mythos}.
For applications in sensitive and higly regulated domains such as medical, legal, or financial NLP,
a prediction that cannot be traced to an explicit pattern cannot be certified. For many real-world
NLP tasks, especially those with recurring lexical or structural patterns, explicit rules remain
attractive: cheap to run, deterministic, easy to version, and straightforward to inspect.  The main
problem of rules is not their accuracy; it is that writing and maintaining them by hand is
labor-intensive and requires substantial domain
expertise~\citep{chiticariu2013rule,shnarch2017grasp,kovacs2022potato}.

RuleChef addresses this gap by using LLMs for \emph{rule learning}: instead of using neural models
at inference time, their role is to translate supervision into executable symbolic rules. Given a
task specification and supervision signal $S$, RuleChef synthesizes a rule set $R = \{r_1, \ldots,
r_k\}$, evaluates it on a held-out development split, clusters the failures, and prompts the LLM to
improve the rules. A patch is accepted only if it improves quality on the held-out split, and each
surviving rule records the precision it reached there. Rules can also be reviewed by human domain
experts, and their feedback as well as explicit corrections can also be used to prompt LLMs to make
updates.

The main contributions of this paper are the following:
\begin{itemize}
    \setlength{\itemsep}{0pt}
    \item A system for synthesizing executable rules from task supervision (examples, corrections,
        free-text feedback, or observed model behavior), separating learning-time LLM use from
        inference-time deterministic execution.
    \item A refinement loop that iteratively improves a rule system, measures the impact of changes
        on held-out data and resolves conflicts between rules by their held-out precision.
    \item Evaluation on two tasks, comparing learned rules against prompting the same LLM and
        against a dedicated neural extractor, with an ablation study isolating the pipeline's
        contribution over one-shot rule prompting.
    \item Additional experiments on the human-in-the-loop repair process and the unsuperviesd observation mode that learns rules
        from the behavior of an external model.
\end{itemize}

The remainder of this paper is structured as follows.
Section~\ref{sec:related} reviews related work on rule synthesis, weak supervision, and LLM-based program generation.
Section~\ref{sec:system} describes the RuleChef framework.
Section~\ref{sec:setup} presents our experimental setup.
Section~\ref{sec:results} reports preliminary evaluation results on a simple intent classification
dataset and a more challenging NER task.
Section~\ref{sec:discussion} discusses key findings and presents opportunities for future work.

\section{Related Work}
\label{sec:related}

In this section we review prior work related to RuleChef across four areas: automatic rule and
regex synthesis, interactive rule systems, weak supervision, and LLM-based code and rule
generation. Automatic regex synthesis from examples has been studied with evolutionary
methods~\citep{bartoli2016inference,bartoli2017active}, neural sequence-to-sequence
models~\citep{locascio2016neural,zhong2018semregex}, and systems that combine natural language
descriptions with positive and negative examples~\citep{chen2020regel,li2021transregex}.  These
approaches produce a single rule or small rule set from examples, but do not include iterative
validation or a human-in-the-loop component.

Interactive rule systems such as HEIDL~\citep{sen2019heidl},
GrASP~\citep{shnarch2017grasp,lertvittayakumjorn2021grasp}, Odinson~\citep{valenzuela2020odinson},
and POTATO~\citep{kovacs2022potato} help users build patterns over texts as well as over
graph-based representations of semantics and syntax (e.g. Abstract Meaning Representations~\citep{banarescu2013amr} and Universal Dependencies~\citep{nivre2018ud})), but still require substantial manual authoring.
RuleChef reduces this manual effort by creating an interface between various sources of supervision
(examples, corrections, feedback on rules) and the LLM performing updates to the rule system.

Weak supervision systems such as Snorkel~\citep{ratner2017snorkel} and Snuba~\citep{varma2019snuba}
use labeling functions to create training labels, and recent work also uses LLM prompts as labeling
functions~\citep{smith2023language,yu2023alfred}.  Unlike these systems, RuleChef makes the rules
the final model rather than a source of noisy labels for a downstream classifier. LLMs have also
been used to synthesize executable programs from examples, including
Evaporate-Code+~\citep{arora2023evaporate} and Hypothesis Search~\citep{wang2023hypothesis},
broader surveys discuss LLM-based rule and hypothesis
generation~\citep{adrian2024llm,survey2025hypothesis}.  RuleChef can also be viewed as the symbolic
alternative for LLM knowledge
distillation~\citep{west2022symbolic,zhou2024universalner,hsieh2023distilling} that maximizes
explainability and minimizes inference cost. For the two tasks evaluated here, the relevant neural baselines are GLiNER~\citep{zaratiana2024gliner}
and its schema-driven successor GLiNER2~\citep{zaratiana2025gliner2} for zero-shot extraction of
text spans, and
dual sentence encoders for intent detection~\citep{casanueva2020efficient,zhang2021cpft}.

\section{The RuleChef Framework}
\label{sec:system}

\begin{figure*}[t]
\centering
\begin{adjustbox}{max width=\textwidth}
\begin{tikzpicture}[
  font=\small,
  box/.style={draw, rounded corners=2pt, align=center, inner sep=4pt, minimum height=2.1em},
  llm/.style={box, fill=orange!15},
  det/.style={box, fill=blue!8},
  lbl/.style={font=\footnotesize\itshape, text=gray},
  arr/.style={-{Stealth[length=2mm]}, thick},
  node distance=4mm and 6mm
]
\node[det] (sup) {supervision\\ \footnotesize examples, corrections,\\ \footnotesize feedback, observations};
\node[llm, right=of sup] (syn) {synthesize\\ \footnotesize per class, grex hints};
\node[det, right=of syn] (eval) {evaluate\\ \footnotesize on dev split};
\node[det, right=of eval] (clu) {cluster\\ \footnotesize failure modes};
\node[llm, right=of clu] (patch) {patch\\ \footnotesize + critic, audit};
\node[det, right=of patch] (acc) {accept iff\\ \footnotesize dev F1 improves};
\node[det, below=7mm of acc] (stamp) {stamp validated\\ \footnotesize precision, support};
\node[det, left=14mm of stamp] (exec) {\textbf{inference:} rules only\\ $\approx$1\,ms/doc};

\draw[arr] (sup) -- (syn);
\draw[arr] (syn) -- (eval);
\draw[arr] (eval) -- (clu);
\draw[arr] (clu) -- (patch);
\draw[arr] (patch) -- (acc);
\draw[arr] (acc.north) to[out=120,in=60] node[above, lbl] {iterate} (eval.north);
\draw[arr] (acc) -- (stamp);
\draw[arr] (stamp) -- (exec);

\begin{scope}[on background layer]
\node[fit=(syn)(acc), draw=gray!60, dashed, rounded corners, inner sep=6pt,
      label={[lbl]above:learning time (LLM involved)}] {};
\end{scope}
\end{tikzpicture}
\end{adjustbox}
\caption{The RuleChef pipeline. Orange components call the LLM; blue components are deterministic. The LLM proposes rules and patches, but acceptance is decided by quality measured on a held-out development split, and each rule records its precision there. At inference time only the rules run.}
\label{fig:pipeline}
\end{figure*}
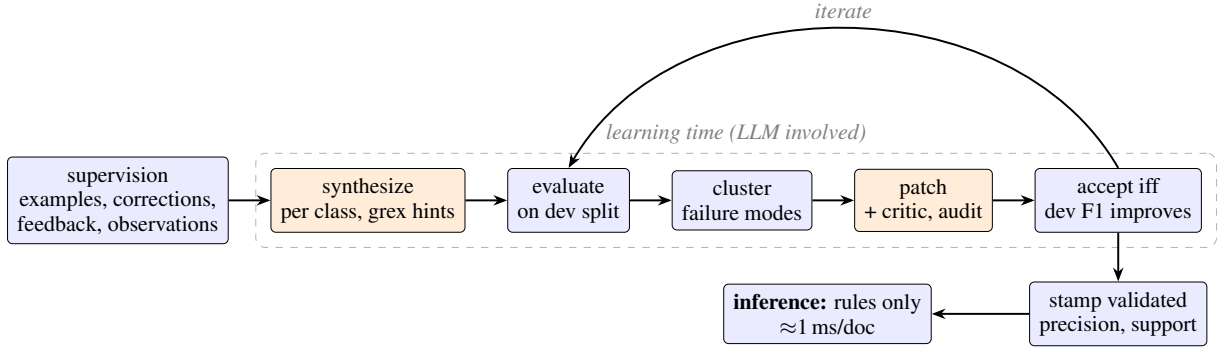

In this section we describe the main functionalities of the RuleChef framework, including task
definition, rule synthesis, the refinement loop, conflict resolution, and the observation mode that
allows autonomous operation in the presence of a pre-existing model. A high-level overview is presented in Figure~\ref{fig:pipeline}.
We draw the examples in this section from the Text Anonymization Benchmark dataset (TAB, see Section~\ref{sec:setup}), our main testbed. The prompt templates corresponding to the various rule learning strategies described here are presented in Appendix~\ref{app:prompts}.

\subsection{Rule Synthesis}
\label{sec:synthesis}

We structure the synthesis prompt in four sections: (1)~the task definition with input/output
schemas, (2)~sampled training examples and corrections, (3)~data evidence with pattern suggestions,
and (4)~format instructions with response schema.  For multi-class tasks, we generate one prompt
per label with positive examples and counter-examples from other classes, preventing cross-class
interference.

The task definition specifies input/output schemas and one of four task types: \textit{Classification}, \textit{NER},
\textit{Extraction} (untyped spans), or \textit{Transformation} (structured output). Rules can be
requested in one of three formats, including regular expressions (regexes), spaCy rules that rely
on the output of part-of-speech tagging and syntactic parsing, and arbitrary Python code. While
each of these three formats is supported by RuleChef, the examples and evaluation in this paper focus only on regex patterns.
Each generated rule is stored along with metadata that specifies its priority and measures its
precision and match-count over the held-out development set.

By default the data evidence provided to the LLM consists of labeled examples, but RuleChef can
also augment this with regex suggestions obtained from \texttt{grex}~\citep{grex}, a regex generator that derives
structural patterns from example strings. Such patterns serve as hints, not
constraints, helping the LLM identify structural regularities without overfitting.
We validate each generated rule before acceptance: regex patterns must compile, output templates
must match the task schema, and patterns that match arbitrary text are rejected by probing them
against generic strings.

\subsection{Refinement}
\label{sec:refinement}

After initial synthesis, a refinement loop evaluates the current rules, identifies failures, and
generates \emph{patch rules} targeting missed or misclassified inputs. Data is split into training and development portions
(explicit user corrections always stay in train, since they are the
highest-value signal for patching), and the synthesis step for rule patching can only access training
data, the dev split is used to decide whether a newly generated rule is accepted. A patch is kept only if held-out F1 does not degrade or
precision improves. This filter prevents memorization: as we show in Section~\ref{sec:ablation}, the same loop without holdout
acceptance drifts toward patterns that fit the failures it was shown rather than ones that
generalize.

With hundreds of training documents, the failures of an intermediate rule set number in the
thousands, far more than fit in a patch prompt.  RuleChef clusters failures by their
signature. For NER-type tasks, failure modes include \textit{missed span}, \textit{spurious span}, and  \textit{wrong type}.
The LLM is presented with the full distribution of failure modes plus a sample of failed instances
from each cluster. The patch prompt also carries the current rule set with per-rule metrics and all
accumulated feedback, an example is shown in Figure~\ref{fig:refinement}. The LLM can modify
existing rules, add new ones, or delete overly broad rules when providing narrower replacements.

\begin{figure}
\noindent\small
\begin{adjustbox}{max width=\columnwidth}
\begin{tabular}{@{}p{7.6cm}@{}}
\texttt{Rule: "case\_and\_echr\_numbers", P=86\% on dev} \\
\texttt{~~failure\_mode: CODE missed\_span (84 cases),} \\
\texttt{~~~~e.g. "nos. 6210/73 and 6877/75"}
\end{tabular}
\end{adjustbox}
    \caption{Example input rule for the refinement step. The rule is referred to by its name,
    followed by its dev-set precision, failure type, and number of matches}
\label{fig:refinement}
\end{figure}

For streaming or batch-wise data, RuleChef can skip the initial synthesis step and update an existing ruleset
directly: we commit new examples, corrections, and feedback, evaluate the current rules, and use
the resulting failures to drive patch synthesis, preserving the rest of the ruleset.  This is the
mechanism behind the human-feedback experiment described in Section~\ref{sec:hitl}.

\paragraph{Agentic coordination.}
The loop can be driven by a fixed schedule or by an LLM that reasons about the current state.  The
\emph{simple coordinator} runs a fixed number of iterations; the \emph{agentic coordinator} instead
reads the per-class metrics after each iteration to steer the next patch, runs a periodic
\emph{critic} that adds rule-level feedback through the same channel as a human, and runs a
periodic \emph{audit} that merges redundant rules or removes ineffective ones, reverting any change
that lowers measured quality. Appendix~\ref{app:prompts} shows the complete prompt templates for
each agent, Section~\ref{sec:ablation} measures their effect.

\subsection{Conflict Resolution and Pruning}
\label{sec:trust}

When learning concludes, we run each rule alone on the development split and record its precision
there along with the number of matches.  When two rules overlap or disagree on a span, the
executor keeps the match from the higher-priority rule, breaking ties based on dev-set
precision.
Low-support estimates are discounted by a Wilson lower bound~\citep{wilson1927probable}, so a rule
right twice out of twice does not outrank one right 95 times out of 100.  A leave-one-out pass
measures each rule's marginal contribution to ensemble F1 and drops rules whose removal does not
hurt it.

Measuring precision on data the rule was not tuned on also separates good rules from memorized
ones.  On the TAB data, the rule \texttt{case\_and\_echr\_numbers}, corresponding to the regex \verb!(?:no\.?\s*)(\d{4,6}/\d{2,4})!, achieves 0.86 precision with 22 true positive matches on the development set.
The overly generic rule \textsc{Quantity}, by contrast, matched only two development spans at precision 1.00, which
the Wilson bound rightly discounts: the rule also matched fraction-shaped case numbers such as
\emph{1432/03} (5.7 F1 on test) before the feedback repair step (see Section~\ref{sec:hitl}).  Rules that
memorized training lexicons show up the same way, with low development-set precision
(see Section~\ref{sec:qualitative}).

\paragraph{Observation mode.}
\label{sec:observation}

For settings where an LLM already handles production traffic, RuleChef can use the model's behavior
as a supervision source rather than requiring labels upfront.  Each LLM call is treated as a
training example, and RuleChef periodically synthesizes rules from the accumulated observations
using the standard pipeline. Matched queries are then routed to rules instead of the LLM. A deployment can delegate only to
rules whose measured precision exceeds a threshold, leaving the rest to the LLM. This mode also
supports \textit{task discovery}, where RuleChef prompts an LLM to discover the full task specification
based on the input-output examples. This mode allows the system to generate an initial rule system
based on an external black box model without human input.

\section{Experimental Setup}
\label{sec:setup}

We provide preliminary evaluation of the RuleChef methodology introduced in this paper
via experiments on two datasets that represent two of the most common NLP tasks, Named Entity
Recoginition and text classification. In this section we describe the datasets and the experimental
setup, results follow in Section~\ref{sec:results}.

\subsection{TAB: Anonymization of Court Decisions}

Our primary experiment uses the Text Anonymization Benchmark~\citep[TAB;][]{pilan2022tab}, a corpus of 1{,}268 decisions
of the European Court of Human Rights, with human annotation marking all text spans that represent
personal information.  We use TAB's eight official entity types unchanged: \textsc{Person}, \textsc{Code}
(case numbers, phone numbers, license plates, etc.), \textsc{Datetime}, \textsc{Quantity}, \textsc{Org}, \textsc{Loc},
\textsc{Dem} (demographic attributes such as nationality or profession), and \textsc{Misc}.  For
analysis we additionally group them by what governs their surface form: \emph{format types}
(\textsc{Code}, \textsc{Datetime}, \textsc{Quantity}), whose mentions follow formal patterns, and
\emph{semantic types} (\textsc{Person}, \textsc{Org}, \textsc{Loc}, \textsc{Dem}, \textsc{Misc}),
defined by meaning rather than form.

Since full documents are too long to fit into the context window of LLMs for rule learning, we
segment them into chunks of at most 600 characters. For the experiments described here we generate a sample of 1{,}000 training
and 600 test chunks. Rule systems generated by RuleChef are evaluated on this test set as well as 
on the official test split, which contains 127 full,
unchunked documents. \emph{RuleChef} uses Kimi-K2.6 as the rule-writing LLM with the agentic coordinator, three refinement iterations, and a 20\% development holdout; at inference time only the learned rules run.
Our main baseline for comparison is \emph{GLiNER2}~\citep{zaratiana2025gliner2}, a 205M-parameter schema-driven extractor.

\subsection{Banking77: Intent Classification and Observation Mode}

The text classification experiments use Banking77~\citep{casanueva2020efficient}, a dataset of over
13k customre service queries classified by user intent into 77 categories that include classes that
are easily detected via keywords (such as \texttt{exchange\_rate}) but also more challenging ones such as\texttt{beneficiary\_not\_allowed}.
A set of 200 examples spanning 25 intent classes is used as the held-out test set in our experiments.
Given the relative simplicity of this task, our baseline for comparison is based on directly
prompting the same LLM that we use for rule generation (Kimi-K2-Instruct). This approach achieves
over 98\% accuracy on this test set and is also used to evaluate RuleChef in Observation mode,
where system-generated labels provide the only source of supervision for rule generation. At each
iteration we measure
rule coverage, precision, and the fraction of subsequent LLM calls replaced.
As context, fine-tuned dual-encoder and contrastive models reach 86--87\% accuracy on the full 77-class task with 10 shots per class~\citep{zhang2021cpft}.

\section{Results}
\label{sec:results}

In this section we report results on the two tasks introduced in Section~\ref{sec:setup}.
Sections~\ref{sec:capability}--\ref{sec:hitl} evaluate on TAB, comparing RuleChef-generated rules
against published baselines and measuring the impact of the feedback-repair process.
Section~\ref{sec:qualitative} provides qualitative analysis of the rules learned on the TAB
dataset.  Section~\ref{sec:banking} uses the Banking77 dataset, measures performance against
few-shot classification approaches, and evaluates  RuleChef's observation mode.

\subsection{Rule system performance}
\label{sec:capability}

\begin{table}[htbp]
\centering
\footnotesize
\setlength{\tabcolsep}{3pt}
\begin{adjustbox}{max width=\columnwidth}
\begin{tabular}{lccccccc}
\toprule
 & \multicolumn{3}{c}{\textsc{Format}} & \multicolumn{3}{c}{\textsc{Semantic}} & \\
\cmidrule(lr){2-4}\cmidrule(lr){5-7}
System & P & R & F1 & P & R & F1 & ms/doc \\
\midrule
RuleChef (rules only) & \textbf{89.1} & 70.5 & \textbf{78.7} & \textbf{75.7} & 34.9 & 47.8 & \textbf{0.6} \\
LLM prompting & 65.1 & \textbf{87.9} & 74.8 & 54.6 & \textbf{60.5} & \textbf{57.4} & $\approx$1500 \\
GLiNER2 (labels) & 66.5 & 78.0 & 71.8 & 31.7 & 46.2 & 37.6 & 190 \\
GLiNER2 (schema) & 69.6 & 79.9 & 74.4 & 34.9 & 49.0 & 40.8 & 190 \\
\bottomrule
\end{tabular}
\end{adjustbox}
\caption{Results on the 600 test chunks of the TAB dataset. The \textit{Format} group includes the entity types \texttt{Code},
    \texttt{Datetime} and \texttt{Quanitity}, the \textit{Semantic} group includes \texttt{Per,
    Org, Loc, Dem,} and \texttt{Misc}.}
\label{tab:tab-main}
\end{table}

Table~\ref{tab:tab-main} compares the performance of learned rules against direct LLM prompting
and two GLiNER2 baselines.
We observe that learned rules are superior to all other systems on classes that can typically be detected based on the
surface form of the entities, while on the more challenging types their overall performance remains below
that of direct LLM prompting, but they still outperform the GLiNER baselines and also achieve the
highest precision among all systems. Higher precision at the cost of lower recall is the intended
behavior, since RuleChef is designed to generate individual rules with high precision, ensuring that incrementally growing a
rule system by adding additional patterns can eventually lead to a system with high precision and
high recall. A qualitative analysis of learned rules is provided in Section~\ref{sec:qualitative}.
We also measure inference latency for each system to quantify the obvious fact that using
regex-based rules means having virtually zero inference costs. Meanwhile, the rule learning process
involved less than 20 LLM calls and took approx. 12 minutes.

\begin{table}[htbp]
\centering
\footnotesize
\setlength{\tabcolsep}{3pt}
\begin{adjustbox}{max width=\columnwidth}
\begin{tabular}{lcccl}
\toprule
System & P & R$_{\text{all}}$ & R$_{\text{direct}}$ & supervision \\
\midrule
RuleChef (22 rules) & .738 & .719 & .830 & 1{,}000 chunks \\
Longformer$^\dagger$ & \textbf{.836} & \textbf{.919} & --- & fine-tuned, 1{,}013 docs \\
RoBERTa NER$^\dagger$ & .441 & .906 & --- & off-the-shelf \\
\bottomrule
\end{tabular}
\end{adjustbox}
\caption{Official TAB test split (127 full documents), scored with the benchmark's own evaluation script: token-level precision weighted by BERT information content, and mention-level recall over annotated spans (all identifiers / direct identifiers only). $\dagger$:~as reported by \citet{pilan2022tab}.}
\label{tab:official}
\end{table}

We also run the learned ruleset over the 127 full, unchunked documents of the official TAB test
split and score it with the benchmark's own evaluation script, which measures recall over annotated
spans and token-level precision weighted by BERT-estimated information content.
Table~\ref{tab:official} compares results with the two baselines reported under the same protocol
by \citet{pilan2022tab}.
Performance of the 22 patterns learned by RuleChef remains well below that of the fine-tuned Longformer
system on both precision and recall, but is already superior to the off-the-shelf RoBERTa system,
achieving nearly 30 points higher precision and only 19 points lower recall. These results
illustrate that RuleChef can generate fully transparent, inspectable and editable rule systems that
are competitive with off-the-shelf neural models. In application scenarios where transparency,
explainability, or auditability of decisions is critical, such high-precision
systems can serve either as the basis of further development of high-performing rule systems or as
the preferred model in a cascade of systems that also include models with higher recall at the cost
of precision and/or explainability.


\subsection{Ablation}
\label{sec:ablation}

\begin{table}[t]
\centering
\footnotesize
\setlength{\tabcolsep}{3.5pt}
\begin{adjustbox}{max width=\columnwidth}
\begin{tabular}{lcccc}
\toprule
Configuration & \textsc{Fmt} F1 & \textsc{Sem} F1 & rules & calls \\
\midrule
one-shot rule prompting & 74.2 & 7.4 & 31 & 8 \\
+ refinement on train & 63.2 & 38.2 & 38 & 12 \\
+ holdout acceptance & \textbf{83.5} & 44.4 & 36 & 14 \\
+ agentic coord.\ (3 iter.) & 80.9 & 35.9 & 21 & 19 \\
\hspace{2mm}agentic, 8 iterations & 81.7 & \textbf{47.8} & \textbf{9} & 33 \\
\bottomrule
\end{tabular}
\end{adjustbox}
\caption{Ablation on TAB. Each row adds one component; ``calls'' counts all learning-time LLM calls (synthesis, patches, critic, audit). One-shot prompting writes plausible format regexes but is nearly useless on semantic types; the validated refinement loop does the work.}
\label{tab:ablation}
\end{table}

A simple ablation study is performed to understand the impact of RuleChef's iterative learning
process on the quality of the final rule system. In particular this involves isolating the baseline
performance of prompting an LLM to generate a rule system in a single pass. We observe that in such
settings LLMs rely on the task description at least as much as on the initial training examples,
and it is the subsequent iterations guided by additional examples and feedback on failures that
allows LLMs to move from a generic rule system towards one that fits the actual task that is represented
by the training dataset. Table~\ref{tab:ablation} shows the effect of various refinement steps on the TAB dataset.
Note that the row \textit{+agentic coord.} corresponds to the main configuration evaluated in
earlier sections, but the numbers differ (e.g. precision of 80.9 vs. 78.7) due to the non-determinism of RuleChef's process
that is inherent to the iterative prompting of LLMs (see also the section on Limitations).

Once again we observe a stark contrast between the two label groups \textit{Format} (FMT) and
\textit{Semantic} (SEM). On the FMT group, which contains entity types easily identified by
their format (dates, quantities, various identification numbers), a single pass of an LLM can
create a ruleset that is within 5-10 percentage points of the best ones in terms of F-score. In
case of the SEM group containing traditional NE types such as \texttt{PER, LOC,} and
\texttt{ORG} an initial rule-system is practically useless.

Refining the rules by providing more training examples achieves substantial improvement on the SEM
group, while the strong baseline on the FMT group deteriorates. However, the additional step of
filtering new rules based on their performance on the dev set (\textit{holdout acceptance})
substantially increases performance on both sets of entity types. The effect of \textit{agentic
coordination} varies greatly with the number of iterations.  In this simple setup we see that 3
iteration leads to a decrease in performance compared to the strongest system on both groups, while
8 iterations produce a superior system on the more SEM group but cannot restore the original top
performance on the FMT group. The total number of rules accross all entity types is of particular interest.
After 3 iterations of the coordinator, the number of patterns is already reduced to 21 from the 36
of the previous system, and after 8 iterations it contains only 9 rules altogether, which still
achieve top scores on the SEM group and are within 2 points of the best system on the FMT group.

We believe that these results, while preliminary, already give a clear indication
that a guided, iterative rule generation approach can result in robust rule-based systems even for challenging tasks like NER.
Properly understanding the nature of how a rule system evolves as a result of such iterative improvement
strategies and how this process varies across types of language processing tasks will require
considerable further experimentation, including also the qualitative analysis of intermediate rule
systems for each task.

\subsection{Repairing Rules with Feedback}
\label{sec:hitl}

Because RuleChef models are lists of readable rules, its defects can also be addressed explicitly
via human-in-the-loop (HITL) feedback. The next set of experiments measures the impact of this process.
We inspected the learned rules and targeted three issues: the \textsc{Quantity} rule matched
fraction-shaped case numbers like \emph{1432/03}, \textsc{Code} rule missed bare application numbers, and a
\textsc{Person} rule for initials fired on ordinary abbreviations. We attached one sentence of
feedback to each rule through the standard interface, e.g.:

\begin{quote}\itshape\small
``Never match number/number patterns like `1432/03'---those are case numbers, not quantities.''
\end{quote}

\begin{table}[htbp]
\centering
\footnotesize
\setlength{\tabcolsep}{5pt}
\begin{adjustbox}{max width=\columnwidth}
\begin{tabular}{lccc}
\toprule
Type & before F1 & after F1 & $\Delta$ \\
\midrule
\textsc{Quantity} (feedback) & 5.7 & 35.6 & \textbf{+29.8} \\
\textsc{Code} (feedback) & 45.4 & 48.8 & +3.4 \\
\textsc{Person} (feedback) & 53.4 & 54.0 & +0.6 \\
\textsc{Datetime} (untouched) & 85.1 & 85.1 & 0.0 \\
\textsc{Dem} (untouched) & 40.4 & 37.7 & $-$2.7 \\
\bottomrule
\end{tabular}
\end{adjustbox}
    \caption{Results of a single round of human-in-the-loop repair using three sentences of rule-level feedback}
\label{tab:hitl}
\end{table}

\noindent Results are shown in Table~\ref{tab:hitl}. A single repair round (92 seconds, two LLM
calls) substantially improves performance on the \textsc{Quantity} class (from 5.7 to 35.6 F1),
while on the other two criticized classes the performance improves modestly. This workflow of
improving individual rules based on a human's understanding is one that hand-crafted rule systems
always supported and neural models do not.

\subsection{Qualitative Analysis of Learned Rules}
\label{sec:qualitative}

\begin{table}[htbp]
\centering
\footnotesize
\setlength{\tabcolsep}{3pt}
\begin{adjustbox}{max width=\columnwidth}
\begin{tabular}{L{2.1cm}L{3.5cm}cc}
\toprule
Rule name & Pattern (abridged) & dev P & $n$ \\
\midrule
\texttt{single\_date} & \texttt{\textbackslash{}b\textbackslash{}d\{1,2\}\textbackslash{}s+ (?:January|...)} & .95 & 362 \\
\texttt{court\_names} & \texttt{(?:District|Regional| Supreme|...)\textbackslash{}s+Court} & .79 & 118 \\
\texttt{titled\_full\_ name} & \texttt{\textbackslash{}b(?:Mr|Mrs|Dr| Justice)\textbackslash{}.?\textbackslash{}s+[A-Z]...} & .93 & 69 \\
\texttt{republic\_ kingdom\_gov.} & \texttt{(?:Federal Republic of Germany|...)} & .97 & 37 \\
\texttt{specific\_ institutions} & \texttt{(?:BBC Scotland|House of Lords|...)} & .47 & 30 \\
\texttt{case\_and\_ echr\_numbers} & \texttt{(?:no\textbackslash{}.?\textbackslash{}s*) (\textbackslash{}d\{4,6\}/\textbackslash{}d\{2,4\})} & .86 & 22 \\
\texttt{initials\_ with\_period} & \texttt{(?:[A-Z]\textbackslash{}. [\textbackslash{}s-]?)\{1,4\}...} & .35 & 17 \\
\texttt{lives\_in\_ location} & \texttt{(?:lives|resides)\textbackslash{}s+in \textbackslash{}s+([A-Z]...)} & 1.00 & 12 \\
\bottomrule
\end{tabular}
\end{adjustbox}
\caption{Examples of rules learned on the TAB dataset, ranked by the number of matches on the dev set.}
\label{tab:rules}
\end{table}

Table~\ref{tab:rules} shows examples of rules learned on the TAB dataset, together with their
precision and match count on the development set. Descriptive rule names have been generated by the
LLMs. We observe that high-coverage high-precision rules encode task knowledge such as formatting
rules (date formats in the rule \texttt{single\_date}, honorifics in the rule
\texttt{titled\_full\_name}, etc.) and lists of terms (\texttt{court\_names},
\texttt{republic\_kingdom\_gov}). In contrast, the rule \texttt{lives\_in\_location} encodes patterns about the context in which
entities of a certain type occur. We can also observe that some rules encoding lists of entities or
formatting conventions achieve substantially lower precision than others. The list of organizations
\texttt{specific\_institutions} is overly generic, also containing standalone words such as
\textit{Parliament}, and so is the pattern \texttt{initials\_with\_period} that matches
abbreviations that are not person names. These are the kinds of rules that must be the target of
subsequent improvement steps that refine them based on examples of false positives presented to the
LLM.

\subsection{Few-Shot Classification and Observation Mode}
\label{sec:banking}

\begin{table}[htbp]
\centering
\footnotesize
\setlength{\tabcolsep}{4pt}
\begin{adjustbox}{max width=\columnwidth}
\begin{tabular}{lrrrrrr}
\toprule
 & Rules & Cov. & Repl. & P & R & F1 \\
\midrule
LLM prompting & --- & 100 & --- & 98.0 & 98.0 & 98.0 \\
Zero-shot NLI (DeBERTa-v3) & --- & 100 & --- & 96.5 & 96.5 & 96.5 \\
LogReg (MiniLM emb.) & --- & 100 & --- & 95.0 & 95.0 & 95.0 \\
RuleChef (few-shot) & 126 & 62.5 & --- & 97.6 & 61.0 & 75.1 \\
\midrule
\multicolumn{7}{l}{\emph{Observation mode (no labels)}} \\
~~after 10 calls & 14 & 20.5 & 19.5 & 92.7 & 19.0 & 31.5 \\
~~after 25 calls & 24 & 42.5 & 41.0 & 95.3 & 40.5 & 56.8 \\
~~after 50 calls & 40 & 49.5 & 48.0 & 96.0 & 49.5 & 63.6 \\
\bottomrule
\end{tabular}
\end{adjustbox}
\caption{Results on the 200 dev set queries of the Banking77 dataset. Top rows compare rule
    performance to three baselines: simple LLM prompting, a logistic-regression classifier on
    MiniLM sentence embeddings, and the DeBERTa-v3 model
    \texttt{MoritzLaurer/\allowbreak deberta-v3-base-zeroshot-v1.1-all-33}
    used for zero-shot classification, which
    scores each candidate label as an entailment
    hypothesis.
    Bottom rows show the performance RuleChef's \textit{observation mode},
    which discovers class labels and learns rules based on a growing number of input-output
    examples from the top-performing LLM-based approach. Coverage (Cov.) measures the
    fraction of inputs covered by any learned rule.}
    \label{tab:banking}
\end{table}

We use the Banking77 dataset to test RuleChef on a simple 5-way intent classification task. The
results in Table~\ref{tab:banking} show that simple LLM prompting as well as standard supervised
learning approaches achieve precision and recall in the .95-.98 range, and that the rule system
learned by RuleChef is competitive with these methods in terms of precision but not in terms of
recall, a typical outcome for rule-based systems. 
The table alsow shows rule performance in various stages of RuleChef's observation mode, where the
system discovers labels and learns rules based on input-output examples from an external model (the
top-performing LLM-based approach). Here we can see
that as the number of observed examples grows from 10 to 50, precision grows from .93 to .96 and
recall increases from .19 to .50. Regardless of whether or not our method would be able ot
further increase recall without sacrificing precision, a system such as the one learned based on
only 50 examples and without human intervention is already robust enough to be used as the first
tier of a hybrid system, replacing black box inference for half of the inputs, offering a clear
gain in both transparency and computational cost.
\section{Discussion and Future Work}
\label{sec:discussion}

We presented the RuleChef methodology for constructing rule-based systems for various
NLP tasks via iterative refinement based on annotated examples and human feedback. Using a generic
Named Entity Recognition task as our main testbed, we have shown that for classes of entities exhibiting highly regular surface patterns
it is possible to learn fully transparent rule-based systems
that are competitive with standard black box approaches. On the more challenging groups of
entities RuleChef can produce a set of high-precision rules with substantial recall that can form
the basis of subsequent improvements and can function as the preferred model in hybrid systems,
increasing explainability and reducing computational cost.

Our simple ablation study offers a preliminary evaluation of the various strategies for iterative
learning implemented by RuleChef. In addition to the process of introducing training examples
gradually and only accepting refinements that improve overall performance, we also showcase the
capabilities of the \textit{agentic coordination} feature, which allows an external model to
orchestrate learning steps based on observed rule performance and perceived rule quality.
RuleChef is released as open-source software under an Apache 2.0
license\footnote{\url{https://github.com/KRLabsOrg/rulechef}},
including all code necessary to reproduce the
experimental results presented in this paper.

RuleChef is a generic framework implementing a variety of approaches to rule learning for NLP, and
the empricial results presented in this paper are strictly preliminary. We expect that the utility
of each RuleChef feature will vary greatly across domain, genres, and datasets. The main approaches
described in Section~\ref{sec:system} would each benefit from in-depth experimental evaluation.
The quality of fule synthesis via LLMs depends on the choice of prompts and their ability to
separate training signals from pre-existing model bias. Iterative improvement by introducing
additional training examples depends heavily on chunking and sampling strategies as well as on the
acceptance criteria for newly introduced rules. The use of direct human feedback should be explored
under various constraints on the form and content of such user input, while agentic coordination
depends on stand-alone models for assessing and criticizing rules, a complex task in its own right.

\section*{Limitations}
\label{sec:limitations}

The broad range of methods introduced and evaluated in this paper introduce a number of
limitations.  Quantitative results presented are based on single runs and fixed splits. Repeated
runs of selected configurations show variations in F1 performance of approx. $\pm$3 points even for
the more regular \texttt{FORMAT} group of entities, and we report metrics from representative
experiments rather than means over multiple runs.  All experiments use English texts as input, and
a single LLM family for rule generation (Kimi-K2). Each main method involving LLM invocations also
introduces the need for the systematic evaluation of how model bias influences the rule learning
process and how this may introduce bias into the final rule systems.

\bibliography{related_work}

\appendix
\input{appendix}

\end{document}

%% file: appendix.tex

\lstdefinestyle{prompt}{%
  basicstyle=\scriptsize\ttfamily,
  breaklines=true,
  breakindent=1.5em,
  breakatwhitespace=false,
  frame=single,
  framesep=3pt,
  columns=flexible,
  keepspaces=true,
  showstringspaces=false,
  backgroundcolor=\color{gray!7},
  morecomment=[l]{\#\#},
  commentstyle=\itshape\color{gray!70},
}

\section{Prompt Templates}
\label{app:prompts}

RuleChef uses large language models exclusively at learning time, issuing
calls through eleven distinct prompt templates that span every phase of rule
synthesis, refinement, agentic coordination, and observation.
The prompts fall into four functional groups.

\subsection{Learning prompts} The \emph{synthesis prompt} (Figure~\ref{fig:prompt-synthesis}) is the primary
learning call: it is issued once at the start of a run to generate an initial
rule set from the full training dataset.  For multi-class tasks, the
\emph{per-class synthesis prompt} (Figure~\ref{fig:prompt-perclass}) is used
instead, producing one focused call per label with its positive examples and
counter-examples from other classes, preventing cross-class interference.
The \emph{patch prompt} (Figure~\ref{fig:prompt-patch}) is issued once per
refinement iteration to incrementally update the rule set by fixing observed
failures; it receives the current rule set with per-rule metrics alongside a
clustered, sampled set of errors.  Six variants of the patch prompt (from
richest to most compact) are pre-built; RuleChef selects the longest one
fitting the context window.

\begin{figure*}[ht]
\begin{lstlisting}[style=prompt,basicstyle=\tiny\ttfamily]
Task: {task.name}
Description: {task.description}

Input schema: {task.input_schema}
Output schema:
{task.output_schema}
[AVAILABLE ENTITY TYPES: {labels}         ## only for NER / classification tasks
Rules MUST use one of these types in output_template.]

CORRECTIONS (Learn from failures - {n} shown):   ## omitted when dataset has no corrections

Input: {json(correction.input)}
Got (WRONG): {json(correction.model_output)}
Expected (CORRECT): {json(correction.expected_output)}
[Feedback: {correction.feedback}]               ## included only when per-correction feedback is set

TRAINING EXAMPLES ({n} shown):

Input: {json(example.input)}
Output: {json(example.expected_output)}

{data_evidence}    ## task-type-specific section: entity counts, grex-derived regex hints,
                   ## class distributions; omitted when evidence is empty

[USER FEEDBACK:                                ## optional; omitted when no feedback is attached
- {feedback_item}]

[EXISTING RULES ({n} current):                ## optional; included on re-synthesis when rules exist
- {rule.name} (priority {rule.priority}, success: {success_rate})
  Format: {rule.format}
  Pattern: {rule.content[:100]}...
  Confidence: {rule.confidence}
CONSIDER:
- Refine existing high-performing rules
- Fix or replace low-performing rules
- Keep rules that work well
- Add new rules for uncovered patterns]

YOUR TASK:
{Update and refine | Synthesize a complete} ruleset (max {max_rules} rules) that:
1. Handles all corrections correctly (CRITICAL - these show failure modes)
2. Works on all examples
3. Respects user feedback
4. Is general and minimal (avoid redundant rules)

WHAT MAKES A GOOD RULE:
- PRECISION OVER RECALL: A rule that matches 10 things correctly beats one that matches 100 with 20 wrong. Never sacrifice precision for recall. Missing a match is fixable later; a wrong match poisons results.
- GENERALIZE, DON'T MEMORIZE: Rules run on unseen text. Match the *structure* of values, not the exact strings you see in training data.
- USE CONTEXT: The same string can mean different things. Match surrounding words/punctuation and use capture groups ($1) to extract just the value.
- ONE RULE, ONE PATTERN: Multiple focused rules beat one giant regex. Each rule should have a clear, testable reason to exist.
- THINK ADVERSARIALLY: Before writing a rule, ask "what else could this match?" If the answer isn't empty, narrow the pattern or add context.

RULES CAN BE:
[- Regex patterns (for structured extraction)]    ## each line included only if the format is allowed
[- Python code (for complex logic)]
[- spaCy token matcher patterns (for linguistic/NLP patterns)]

IMPORTANT: You must ONLY use the allowed formats listed above.
[IMPORTANT: For CODE rules, write standard multi-line Python functions with proper indentation.]
[IMPORTANT: spaCy rules must be valid JSON arrays of token dicts (spaCy Matcher format).]
[IMPORTANT: spaCy NER is disabled. Do NOT use ENT_TYPE or ENT_ID in spaCy patterns.]

[REGEX SYNTAX REFERENCE:                      ## included when REGEX format is allowed
- \b word boundaries, (?:...) non-capturing groups, (a|b|c) alternation
- [A-Z] [a-z] \d \s character classes, +*?{n,m} quantifiers
- (?=...) lookahead, (?<=...) lookbehind -- match context without consuming it CAPTURE GROUPS ($1) vs FULL MATCH ($0): Use $0 when the entire match IS the value; use $1 when context is needed. With "text": "$1", $start/$end auto-adjust to the group position.
GOOD vs BAD PATTERNS:
  BAD:  "\d+"   -- matches every number; most will be wrong type
  GOOD: "(\d+)\s*(?:kg|lbs|oz)"  -- only numbers followed by weight units
  BAD:  "Alice|Bob|Tokyo"  -- memorizes training strings, misses unseen values
  BAD:  "[A-Z][a-z]+"  -- matches any capitalized word (too broad)
  GOOD: "(?:in|from|at|near)\s+([A-Z][a-z]+(?:\s+[A-Z][a-z]+)*)"  -- locations after prepositions
  Prefer structural patterns over memorized alternations.]

[SPACY TECHNIQUES:                            ## included when SPACY format is allowed
- Use POS tags (PROPN, NOUN, VERB), DEP labels (nsubj, dobj, pobj), SHAPE for structural patterns.
- Use IN for value sets; OP for quantifiers ("?" optional, "+" one or more).]

(...)  ## instructions on return format omitted due to space constraints

[{code_or_spacy_format_examples}]            ## task-type-specific format examples, when applicable

Focus on learning from CORRECTIONS - they show exactly what went wrong!

IMPORTANT: Return ONLY valid JSON. Ensure:
- All strings use double quotes and are properly escaped
- All braces and brackets are balanced
- No trailing commas
- Response is complete (not truncated)
\end{lstlisting}
\caption{Rule synthesis prompt.
  Issued once at the start of a learning run to generate an initial rule set
  from the full training dataset.  Optional sections (shown in square brackets)
  are included only when the corresponding data or configuration is present.
  Variable slots are shown as \texttt{\{name\}}; lines beginning with
  \texttt{\#\#} are annotations in this figure, not part of the prompt.
  The data evidence section is filled with task-type-specific pattern statistics
  and \texttt{grex}-derived regex hints.}
\label{fig:prompt-synthesis}
\end{figure*}

\begin{figure*}[ht]
\begin{lstlisting}[style=prompt]
Task: {task.name}
Description: {task.description}

Input schema: {task.input_schema}
Output schema:
{task.output_schema}
[AVAILABLE ENTITY TYPES: {labels}
Rules MUST use one of these types in output_template.]

FOCUS: Generate rules for class '{target_class}' ONLY.
##     (or: "to extract the '{target_class}' field" for TRANSFORMATION tasks)

POSITIVE EXAMPLES for '{target_class}' ({n} total):

Input: {json(example.input)}
Output: {json(example.expected_output)}

[COUNTER-EXAMPLES (these are NOT '{target_class}' -- your rules must NOT match these):
                                             ## omitted for TRANSFORMATION tasks
Input: {json(counter_example.input)}
Label: {json(counter_example.expected_output)}]

{data_evidence}

{format_instructions}

WHAT MAKES A GOOD RULE:                     ## RULE_QUALITY_GUIDE constant (full text in Figure 3)
[...]

INSTRUCTIONS:
- Generate up to {max_rules} rules that match examples of '{target_class}'.
- Rules should generalize to unseen text, not just memorize the examples shown.
- Use structural patterns with word boundaries that generalize to unseen text.
[- Rules must NOT match the counter-examples shown above.]  ## omitted for TRANSFORMATION

{response_schema}

{format_examples}

Focus on learning from CORRECTIONS - they show exactly what went wrong!

IMPORTANT: Return ONLY valid JSON. Ensure:
- All strings use double quotes and are properly escaped
- All braces and brackets are balanced
- No trailing commas
- Response is complete (not truncated)
\end{lstlisting}
\caption{Per-class synthesis prompt.
  Used in place of the synthesis prompt for multi-class tasks: one call is
  issued per class label, showing only that class's positive examples and
  counter-examples from other classes.  Prevents cross-class interference in
  the generated rules.  Not used for binary or untyped extraction tasks.}
\label{fig:prompt-perclass}
\end{figure*}

\begin{figure*}[ht]
\begin{lstlisting}[style=prompt]
You are updating an existing rule-based extractor. Do NOT rewrite good rules; add or adjust
only what is needed.

Task: {task.name}
Description: {task.description}

Input schema: {task.input_schema}
Output schema: {task.output_schema}
[AVAILABLE ENTITY TYPES: {labels}
Rules MUST use one of these types in output_template.]

{data_evidence}                              ## omitted in compact variants to save tokens

[USER GUIDANCE (task-level feedback):
- {feedback_item}]

[COORDINATOR GUIDANCE (prioritize this):
{guidance}]                                  ## injected by AgenticCoordinator.guide_refinement()

[PER-CLASS METRICS (current performance):
  {class}: P={p} R={r} F1={f1} (TP={tp} FP={fp} FN={fn})]

CURRENT RULES (full details, note any user_feedback on specific rules):
{json(rules_detail)}        ## each entry: name, format, pattern, metrics, user_feedback

[OTHER CURRENT RULES (compact index; avoid duplicating these names/labels):
{json(compact_rules)}]      ## non-relevant rules shown compact in the "relevant" variant

{failure_summary}           ## mode-based count of all failures, not just the sampled set below
FAILURES TO FIX (sampled, corrections are high priority):
{json(failure_snippets)}    ## up to 20/10/5 items depending on variant; corrections: is_correction=true

[FALSE POSITIVES (rules are incorrectly matching these -- tighten the responsible rules):
  Predicted "{predicted_text}" as {predicted_type}
    -- {not an entity | should be {correct_type}}]

Instructions:
- Add, tweak, or DELETE rules to fix the shown failures and reduce false positives.
- Pay close attention to user_feedback on rules AND task-level USER GUIDANCE -- these are direct
  instructions from the user and MUST be addressed even if there are no failures.
- If a rule has user_feedback, modify or replace that rule to address the feedback.
- IMPORTANT: When updating an existing rule, you MUST reuse the EXACT same "name" as the original.
  Do NOT add suffixes like "_fixed", "_v2", "_updated", etc. The merge system uses name-matching to
  replace the old version -- a different name creates a duplicate instead of replacing.
- If a rule is fundamentally too broad (FP >> TP) and you're providing better, narrower
  replacements, list the old rule's exact name in "deleted_rules". Only delete if you're
  providing replacements in "rules".
- If a rule has high false positives, TIGHTEN its pattern or DELETE it and add narrower
  replacements. Adding context or narrowing the match is better than piling on new rules.
- Use structural patterns with word boundaries that generalize to unseen text.
- Keep total new/updated rules <= {max_rules}.
- Use formats: {allowed_formats}
- Avoid touching unrelated behaviors.

WHAT MAKES A GOOD RULE:                     ## RULE_QUALITY_GUIDE constant (full text in Figure 3)
[...]

{format_instructions}

{response_schema}
\end{lstlisting}
\caption{Patch prompt.
  Issued once per refinement iteration to incrementally fix failures.
  Six variants are pre-built and differ in the number of failure snippets
  shown (20, 10, or 5), whether the data evidence section is included, and
  whether non-relevant rules are shown in compact or full form.  RuleChef
  selects the longest variant that fits the context window.  Lines beginning
  with \texttt{\#\#} are annotations in this figure only.}
\label{fig:prompt-patch}
\end{figure*}

\subsection{Agentic coordinator prompts}
When the agentic coordinator is enabled, four additional prompts drive
autonomous decision-making.  After each refinement iteration the
\emph{refinement coordinator prompt}
(Figure~\ref{fig:prompt-coordinator}) analyses per-class metrics and returns
strategic guidance for the next patch call.  Periodically, the \emph{rule
critic prompt} (Figure~\ref{fig:prompt-critic}) performs a holistic review of
the rule set and produces actionable per-rule feedback, which re-enters the
system through the standard feedback interface and is visible in subsequent
patch prompts.  The \emph{rule auditor prompt}
(Figure~\ref{fig:prompt-auditor}) compares rules for overlap or redundancy
and proposes merge, remove, or tighten actions; any structural change is
verified on the held-out split before being accepted.  When RuleChef is
configured to learn from a live data stream, the \emph{learning trigger
decision prompt} (Figure~\ref{fig:prompt-trigger}) decides whether accumulated
new examples warrant starting a retraining loop.

\begin{figure}[ht]
\begin{lstlisting}[style=prompt]
You are the Refinement Coordinator for a rule learning
system. After each refinement iteration, you analyze
per-class performance and guide the next patch.

ITERATION: {iteration + 1}/{max_iterations}
OVERALL: accuracy={exact_match}, micro_F1={micro_f1},
         macro_F1={macro_f1}

PER-CLASS PERFORMANCE (sorted worst to best):
  {class}: F1={f1} P={p} R={r} (TP={tp} FP={fp} FN={fn})
  [...]

Return JSON:
{
  "focus_classes": ["class names needing most improvement"],
  "guidance": "Specific advice for the rule generator --
    which classes to prioritize, what patterns to try,
    what to avoid. 2-3 sentences max.",
  "should_continue": boolean
}
\end{lstlisting}
\caption{Refinement coordinator prompt.
  Issued after each iteration when the agentic coordinator is enabled.
  The returned \texttt{guidance} string is injected verbatim into the next
  patch prompt as coordinator guidance; \texttt{should\_continue}\,=\,false
  terminates the refinement loop early.}
\label{fig:prompt-coordinator}
\end{figure}

\begin{figure*}[ht]
\begin{lstlisting}[style=prompt]
You are a Rule Auditor for a rule-based extraction/classification system.

Analyze these {n} rules and their per-rule metrics.

RULES:
{json(rule_entries)}  ## each entry: id, name, format, pattern[:300], priority,
                      ## output_key, output_template, precision, recall, F1, TP, FP

Your job is to CONSOLIDATE and CLEAN the ruleset.

ACTIONS (in priority order):
1. MERGE: Two+ rules with similar/overlapping patterns targeting the same output/label.
   Combine into one rule. Only merge rules of the same format and same output_template/output_key.
2. REMOVE rules that hurt more than they help:
   - precision=0 AND matches>0 (pure noise -- every match is wrong)
   - false_positives > 2x true_positives (rule causes more harm than good)
   - Memorized exact strings from training data that won't generalize
3. TIGHTEN: If a rule has high FP, return it as a merge-with-self -- same rule_id but narrower
   pattern.

IMPORTANT -- do NOT remove:
- The only rule for a class/label -- even if it looks weak, tighten it instead
- Rules with 0 matches -- the training set may be small, they could help on unseen data

LOOK FOR:
- Near-duplicate rules (same type, similar regex) -- merge them
- Overly broad patterns (like bare \d+ or [A-Z][a-z]+) with high FP -- tighten or remove
- Rules that are subsets of other rules (one pattern already covered by another)

Return JSON:
{
  "analysis": "Brief summary (1-2 sentences)",
  "actions": [
    {"action": "merge", "rule_ids": ["id1", "id2"], "merged_pattern": "new regex",
     "merged_name": "Combined rule name", "reason": "why"},
    {"action": "remove", "rule_ids": ["id"], "reason": "why"}
  ]
}

Return {"analysis": "All rules are useful", "actions": []} if no changes needed.
\end{lstlisting}
\caption{Rule auditor prompt.
  Issued periodically by the agentic coordinator to merge redundant rules,
  remove pure-noise rules, and tighten over-broad patterns.  Any structural
  change proposed here is applied and then re-evaluated on the held-out split;
  changes that lower F1 are reverted.  Only fires when there are at least two
  rules.}
\label{fig:prompt-auditor}
\end{figure*}

\begin{figure*}[ht]
\begin{lstlisting}[style=prompt]
You are an expert Rule Critic acting as a human domain expert. You are reviewing a rule-based
{task.type} system and providing actionable feedback.

TASK: {task.name}
{task.description}
Type: {task.type}
Input schema: {json(task.input_schema)}
Output schema: {json(task.output_schema)}

OVERALL PERFORMANCE:
  micro F1={micro_f1}, P={micro_precision}, R={micro_recall}
  exact_match={exact_match} ({total_docs} documents)

PER-CLASS PERFORMANCE (sorted worst-first):
  {class}: P={p} R={r} F1={f1} (TP={tp} FP={fp} FN={fn})

RULES ({n} total):
  Rule: "{rule.name}" (id={rule.id})
    Format: {rule.format}, Priority: {rule.priority}
    Pattern: {rule.content}
    [Output template: {json(rule.output_template)}]
    [Output key: {rule.output_key}]
    Metrics: P={p} R={r} F1={f1} (TP={tp} FP={fp}, {total} total matches)
    [FP examples from this rule:
      Input: "{input_context[:150]}"
      Rule produced: {json(rule_output[:3])}
      Expected: {json(expected[:3])}]

[FALSE POSITIVES (system-level, {n} examples):
  Predicted "{predicted_text}" as {predicted_type}
    -- {not an entity | should be {correct_type}} (context: "{context[:80]}")]

[MISSED ENTITIES (sample documents with errors):
  Input: "{input_text[:150]}"
  Expected: {json(expected)}
  Got: {json(got)}]

ANALYZE HOLISTICALLY:
1. Which rules cause the most harm and WHY? Show your reasoning.
2. Are there inter-class conflicts? (same text matched by rules for different types)
3. Are priority assignments correct? (higher priority runs first, wins conflicts)
4. What patterns are MISSING for classes with low recall?
5. What would a human regex expert change about these patterns?

PROVIDE FEEDBACK:
- rule_feedback: For EACH problematic rule, provide SPECIFIC, ACTIONABLE advice.
  Bad:  "This rule is too broad" (vague)
  Good: "Narrow \d+ by adding word-boundary context: use (\d+)\s*(?:million|billion) for large
        numbers, and let MONEY/PERCENT rules handle $-prefixed numbers by higher priority"
- task_guidance: Strategic advice about the ENTIRE ruleset.

Return JSON:
{
  "analysis": "1-2 sentence summary of the main issues",
  "rule_feedback": {
    "{rule_id}": "Specific actionable advice..."
  },
  "task_guidance": "Strategic guidance about the full ruleset..."
}
\end{lstlisting}
\caption{Rule critic prompt.
  Issued periodically by the agentic coordinator to perform holistic expert
  review of the full rule set.  The returned \texttt{rule\_feedback} entries
  are attached to the named rules as feedback items;
  \texttt{task\_guidance} is stored as task-level feedback.
  Both re-enter the system through the same interface as human feedback and
  are visible in subsequent patch prompts.}
\label{fig:prompt-critic}
\end{figure*}

\begin{figure}[ht]
\begin{lstlisting}[style=prompt]
You are the Coordinator for a rule learning system.
Decide if we should trigger a retraining loop NOW
based on new data.

STATUS:
- New examples: {n_examples}
- New corrections (high priority): {n_corrections}
- Current rules: {n_rules}

NEW DATA SAMPLES (up to 10):
- [CORRECTION] Input: {json(ex.input)} -> Output: {json(ex.output)}
- [EXAMPLE]    Input: {json(ex.input)} -> Output: {json(ex.output)}

DECISION CRITERIA:
1. TRIGGER if we have corrections (users fixing mistakes).
2. TRIGGER if we have a significant batch of new examples (5+).
3. WAIT if data looks sparse or redundant.

STRATEGIES:
- 'balanced': Standard mix (default)
- 'corrections_first': If we have corrections
- 'diversity': If we have many similar examples
- 'uncertain': If examples look ambiguous

Return JSON:
{
  "should_learn": boolean,
  "strategy": "balanced"|"corrections_first"|
               "diversity"|"uncertain",
  "max_iterations": integer (1-3),
  "reasoning": "Short explanation"
}
\end{lstlisting}
\caption{Learning trigger decision prompt.
  Issued when RuleChef operates in streaming or batch-wise mode and a new
  buffer of data has accumulated.  The coordinator decides whether to start
  a retraining loop immediately or wait for more data, and selects the
  training strategy and iteration budget.}
\label{fig:prompt-trigger}
\end{figure}

\subsection{Observation mode prompts}
In observation mode, where RuleChef bootstraps rules from an existing LLM's
input-output pairs rather than from manually labelled data, two prompts are
used.  The \emph{task discovery prompt}
(Figure~\ref{fig:prompt-discovery}) infers the task schema (type,
input/output fields) from a sample of raw API call logs.  The
\emph{observation mapping prompt} (Figure~\ref{fig:prompt-mapping}) then maps
each batch of up to ten raw observations to the discovered schema, filtering
irrelevant calls.

\begin{figure}[ht]
\begin{lstlisting}[style=prompt]
You are analyzing LLM API calls to discover the
underlying task pattern.

Here are {n} sample LLM interactions:

{obs_text}    ## formatted raw API logs:
              ## system prompt, user message, assistant response

Based on these interactions, identify:
1. What task is being performed? (name and description)
2. What type of task? Choose ONE: extraction, ner,
   classification, transformation
   - extraction: finding text spans (untyped)
   - ner: finding typed entities with labels
   - classification: assigning a label to input text
   - transformation: extracting structured fields from text
3. What are the input fields and their types?
4. What are the output fields and their types?
5. Which input field contains the main text?

Return ONLY valid JSON:
{
  "name": "task_name",
  "description": "one sentence description",
  "type": "classification",
  "input_schema": {"text": "str"},
  "output_schema": {"label": "str"},
  "text_field": "text"
}
\end{lstlisting}
\caption{Task discovery prompt.
  Issued once in observation mode when no task schema is provided upfront.
  Requires a minimum of accumulated raw observations (default: 5) before
  firing.  The discovered schema is used for all subsequent observation
  mapping calls.}
\label{fig:prompt-discovery}
\end{figure}

\begin{figure}[ht]
\begin{lstlisting}[style=prompt]
You are extracting structured training data from LLM
interactions.

Task definition:
  Name: {task.name}
  Description: {task.description}
  Type: {task.type}
  Input schema:  {json(task.input_schema)}
  Output schema: {json(task.output_schema)}

Here are {n} LLM interactions to analyze:

{obs_text}    ## up to 10 raw API call logs per batch

For EACH interaction, determine:
1. Is it relevant to the task above? (relevant: true/false)
2. If relevant, extract the input (matching input_schema
   keys) and output (matching output_schema keys).

Return ONLY a JSON array with exactly {n} objects:
[
  {"relevant": true,  "input": {...}, "output": {...}},
  {"relevant": false, "input": null,  "output": null},
  ...
]
\end{lstlisting}
\caption{Observation mapping prompt.
  Issued once per batch of up to ten raw API call logs.  Filters irrelevant
  calls and extracts structured input-output pairs matching the task schema.
  Mapped examples are added to the training buffer and consumed by subsequent
  synthesis or patch calls.}
\label{fig:prompt-mapping}
\end{figure}

\subsection{Utility prompts}
Two lightweight prompts cover auxiliary functions.  The \emph{synthetic
example generation prompt} (Figure~\ref{fig:prompt-generation}) generates a
single realistic training input on demand, used to augment sparse training
sets.  The \emph{LLM fallback prompt} (Figure~\ref{fig:prompt-fallback}) is
the only prompt active at inference time: it is issued as a last resort when
no rule fires and the executor is configured to delegate to an LLM rather
than abstain.

\begin{figure}[ht]
\begin{lstlisting}[style=prompt]
Generate a realistic training example for this task:

Task: {task.name}
Description: {task.description}
Input schema: {task.input_schema}

Return JSON with input fields only.
Example #{seed + 1}:
\end{lstlisting}
\caption{Synthetic example generation prompt, generates a single realistic training
    input on demand. The \texttt{seed} parameter drives diversity across calls.}
\label{fig:prompt-generation}
\end{figure}

\begin{figure}[ht]
\begin{lstlisting}[style=prompt]
Task: {task.name}
Description: {task.description}

Input: {json(input_data)}

Output schema: {task.output_schema}

Return ONLY valid JSON matching the output schema,
no explanation.
\end{lstlisting}
\caption{LLM fallback execution prompt: the only prompt active at inference time,
  issued only when no rule fires and the executor is not configured to abstain.}
\label{fig:prompt-fallback}
\end{figure}